%% file: neurips_2023.tex
\pdfoutput=1
\documentclass{article}

   \usepackage[nonatbib,preprint]{neurips_2023}

\usepackage[utf8]{inputenc} 
\usepackage[T1]{fontenc}    
\usepackage{hyperref}       
\usepackage{url}            
\usepackage{booktabs}       
\usepackage{amsfonts}       
\usepackage{nicefrac}       
\usepackage{microtype}      
\usepackage{xcolor}         

\usepackage{amsmath,amssymb}
\usepackage{graphicx}
\usepackage{caption}
\usepackage{textcomp}
\usepackage{xcolor}
\usepackage{algorithm}
\usepackage[noend]{algpseudocode}
\usepackage{float} 

\usepackage{mathrsfs}
\usepackage{CJK}

\usepackage{multirow}
\usepackage{subcaption}
\usepackage{enumitem}
\usepackage{array}
\usepackage{colortbl}

\usepackage[square,numbers]{natbib}

\newcommand{\comment}[1]{}

\title{TabGSL: Graph Structure Learning for Tabular Data Prediction}

\author{%
  Jay Chiehen Liao \\
  Institute of Data Science\\
  National Cheng Kung University\\
  \texttt{jay.chiehen@gmail.com} \\
  \And
  Cheng-Te Li \\
  Institute of Data Science\\
  National Cheng Kung University\\
  \texttt{chengte@ncku.edu.tw} \\
}

\begin{document}

\maketitle

\begin{abstract}
This work presents a novel approach to tabular data prediction leveraging graph structure learning and graph neural networks. Despite the prevalence of tabular data in real-world applications, traditional deep learning methods often overlook the potentially valuable associations between data instances. Such associations can offer beneficial insights for classification tasks, as instances may exhibit similar patterns of correlations among features and target labels. This information can be exploited by graph neural networks, necessitating robust graph structures. However, existing studies primarily focus on improving graph structure from noisy data, largely neglecting the possibility of deriving graph structures from tabular data. We present a novel solution, Tabular Graph Structure Learning (TabGSL), to enhance tabular data prediction by simultaneously learning instance correlation and feature interaction within a unified framework. This is achieved through a proposed graph contrastive learning module, along with transformer-based feature extractor and graph neural network. Comprehensive experiments conducted on 30 benchmark tabular datasets demonstrate that TabGSL markedly outperforms both tree-based models and recent deep learning-based tabular models. Visualizations of the learned instance embeddings further substantiate the effectiveness of TabGSL.
\end{abstract}

\input{1-Introduction.tex}

\input{2-RelatedWork.tex}
\input{3-Problem.tex}

\input{4-Method.tex}
\input{5-Experiments.tex}
\input{6-Conclusions.tex}


\section*{Broader Impact}
Graph structure learning for tabular data holds immense potential in broadening the applicability of machine learning within the real-world context. Tabular data, being the most prevalent data type, encompasses a wide variety of domains including medical and financial demographics, predominantly found in database structures. 
Despite the remarkable breakthroughs in deep learning applications for image and text data, a similar level of success hasn't been realized for tabular data. 
This paper focuses on developing a graph structure learning method for tabular data.
Modeling the latent relationships between data instances with graph structure learning exhibits promising performance improvement, which could potentially revolutionize the way we interpret and analyze tabular data. This also can unlock numerous possibilities for enhancing tabular data utilization across diverse areas like transfer learning, distributed learning, and multi-view learning. 
However, it may inadvertently introduce biases and violate privacy norms, presenting challenges that must be addressed with careful consideration.

{
\small
\bibliographystyle{plainnat}
\bibliography{9-References}
}

\input{A-Appendix}








\end{document}

%% file: 1-Introduction.tex
\section{Introduction}\label{sec-intro}

Tabular data is a common data type in the real world \cite{borisov2021deep}. Among various machine learning (ML) algorithms, gradient boosted decision trees (GBDTs) have been one type of the most competitive ML to handle tabular data for many years \cite{gilmer2017neural, huang2020tabtransformer, gorishniy2021revisiting}. Recently, many studies also tried applying deep learning (DL) methods to tabular data to improve prediction performance, which has become popular in academics and the industry \cite{gilmer2017neural, guo2021tabgnn}. Studies found that DL methods designed for other domains are also helpful for tabular data, such as Fi-GNN~\cite{fignn19cikm}, Table2Graph~\cite{table2graph}, and T2G-Former~\cite{t2gformer23}.
In these methods, feature interactions are taken into account. However, associations among instances are usually ignored in these approaches. Some instances may share similar patterns of features correlated with the prediction target, which can be modeled by a graph.

Recently, studies on graphs have flourished in various domains \cite{wu2020GNNsurvey}. Due to its powerful capability to learn latent representations from relational structure, DL methods are applied to modeling graphs and lead to a thriving research topic, graph neural networks (GNNs) \cite{wu2020GNNsurvey, liu2022sublime}. 
GNNs learn latent representations for each node by aggregating information from the node's neighbors on the given graph structure. 
However, graph structures are usually generated from complex systems and thus are inevitably noisy \cite{zhu2021gslsurvey}. Specifically, there might be redundant information in a graph structure. Some connections might be missing or even incorrect \cite{chen2020IDGL}. Without a doubt, these noises on the graph structure are harmful to GNN's performance considerably. Besides, GNN is not available when modeling data without an apparent graph structure, such as tabular data, which is very common in the real world \cite{liu2022sublime}. Such issues lead to studies focusing on \textit{graph structure learning} (GSL), which aims to learn a reliable graph structure with less noise for GNNs. Although there have been GSL methods proposed, they mainly focused on refining an existing graph structure. Not many discussed learning graph structures purely from tabular data, where there is no available graph initially \cite{chen2020IDGL, liu2022sublime, zhu2021gslsurvey}. GNNs can model the latent associations among instances, which cannot be captured by existing tree-based models. To this end, 
learning a credible graph structure from the given tabular dataset is crucial.

Most of the existing DL methods modeling tabular data consider feature interactions. Feature reconstruction, such as VIME~\cite{vime20icml} and TabNet~\cite{arik2021tabnet}, and contrastive learning, such as SubTab~\cite{ucar2021subtab} and SCARF~\cite{bahri2022scarf}, are well-adopted approaches.
However, the latent associations among instances are almost ignored. Thus, patterns of features and the response variable that some similar instances (i.e., neighbors) share cannot be identified and exploited in tabular data learning. 
Data instances can be correlated with each other in terms of their features.
For example, users with similar profiles or online behaviors tend to have similar preferences for ads or items~\cite{rim21,dgenn21,grecplus22}. Patients with similar clinical data or symptoms have a higher potential to suffer from similar diseases~\cite{gct20,medgraph20,hcl22}. To better represent instances for downstream tasks, rather than solely employing each instance's self features, it is crucial to model the correlation between instances. The key idea is to exploit such correlation to learn higher-quality feature representations of instances, i.e., instances with similar labels are close to one another while those with different labels are pushed away from each other in the embedding space. With proper learning of the graph structure that depicts the relationships between instances, GNN would be a good fit to let instances learn to represent each other. That said, we require both GSL and GNNs to model instance association.

Regarding the learning of graph structure for tabular data prediction, three challenges need to be tackled. First, existing GSL methods require a noisy graph as the input for refinement and adjustment, but tabular data contains no graph topology in essence. We need to learn the graph structure from scratch. 
Second, both modeling feature interactions and instance associations are essential in tabular data learning. How to simultaneously consider them both for tabular data in a unified framework is unclear.
Third, in addition to learning the graph structure from tabular data, we further need to jointly train a graph neural network to obtain feature representations of instances for final predictions. 

In this work, we propose a novel graph machine learning model, \underline{\textbf{Tab}}ular \underline{\textbf{G}}raph \underline{\textbf{S}}tructure \underline{\textbf{L}}earning (\textbf{TabGSL}), for tabular data prediction. The key idea is to learn the graph structure from a given tabular dataset so that the latent correlation between instances can be modeled. We propose a novel contrastive learning mechanism to learn the graph structure from tabular data. The intuition is to construct a teacher, which possesses confident knowledge about label knowledge among instances, to guide a student graph learner to continuously adjust the graph structure. Besides, to further capture the interactions between features, we adopt the transformer with tokenized features to produce feature embeddings for GSL. The learning of tabular graph structure is jointly trained with a GNN module to generate final representations of instances.

We summarize the contributions of this work as follows. \textbf{(a)} We revisit the tabular data prediction task from the perspective of graph structure learning and graph neural networks, and highlight the potential of modeling the associations between instances and capturing the interactions between features. \textbf{(b)} We present a new graph-based solution, Tabular Graph Structure Learning (TabGSL), to better perform tabular data prediction by simultaneously learning instance correlation and feature interaction in a united framework. A novel graph contrastive learning module is devised to fulfill the goal. \textbf{(c)} Experiments conducted on 30 benchmark tabular datasets exhibit that the proposed TabGSL significantly and consistently outperforms tree-based models and recent deep learning-based tabular models. Visualization plots also show the effectiveness of the learned instance embeddings.

This paper is organized as below. We review relevant studies in Section~\ref{sec-related}, followed by the description of the problem statement in Section~\ref{sec-problem}. The methods and tasks involved in learning graph structures for tabular data prediction are presented in Section~\ref{sec-method}. We give the evaluation plan and discuss the experimental results in Section~\ref{sec-exp}. Section~\ref{sec-conclude} concludes this work.

%% file: 2-RelatedWork.tex
\section{Related Work}\label{sec-related}

    \begin{table*}
      \caption{\textbf{Comparison of Relevant Studies.} All works belong to deep learning-based tabular data prediction methods. \textbf{Tabular}: whether a method is created for tabular data prediction; \textbf{Graph}: whether a method belongs to a graph-based approach; \textbf{Task}: the downstream prediction task, including Tabular Classification (TC), Node Classification (NC), and Click-Through Rate prediction (CTR); \textbf{FI}: whether a method models feature interactions; \textbf{IA}: whether a method models instance associations; and \textbf{Training} settings include end-to-end (n2n), Unsupervised (Unsup.).}
      \centering
  \begin{tabular}{l | c c c c c c c}
    \toprule
    \textbf{Method} & \textbf{Tabular} & \textbf{Graph} & \textbf{GSL} & \textbf{Task} & \textbf{FI} & \textbf{IA} & \textbf{Training} \\
    \hline
    IDGL \cite{chen2020IDGL} &  & \checkmark & \checkmark & NC &  & \checkmark & n2n \\
    
    SLAPS \cite{Fatemi2021SLAPS} &  & \checkmark & \checkmark & NC &  & \checkmark & Unsup. \\
    
    SUBLIME \cite{liu2022sublime} &  & \checkmark & \checkmark & NC &  & \checkmark & Unsup. \\
    
    DeepFM \cite{guo2017deepfm} & \checkmark &  &  & CTR & \checkmark &  & n2n \\
    
    TabNN \cite{ke2018tabnn} & \checkmark &  &  & TC & \checkmark &  & n2n \\
    
    TabNet \cite{arik2021tabnet} & \checkmark &  &  & TC & \checkmark &  & n2n \\
    
    TabTransformer \cite{huang2020tabtransformer} & \checkmark &  &  & TC & \checkmark &  & n2n \\
    
    FT-Transformer \cite{gorishniy2021revisiting} & \checkmark &  &  & TC & \checkmark &  & n2n \\
    
    NPT \cite{kossen2021self} & \checkmark &  &  & TC & \checkmark & \checkmark & n2n \\
    
    SubTab \cite{ucar2021subtab} & \checkmark &  &  & TC &  &  & Unsup. \\
    
    Regularized DNNs \cite{kadra2021well} & \checkmark &  &  & TC &  &  & n2n \\
    
    TabGSL (\textit{This work}) & \checkmark & \checkmark & \checkmark & NC & \checkmark & \checkmark & n2n \\
    
    \bottomrule
\end{tabular}\label{tab:related_works}
    \end{table*}

\textbf{Deep Learning for Tabular Data.} 
Tree-based models, such as Xgboost \cite{xgboost}, CatBoost \cite{catboost}, and LightGBM \cite{lightgbm}, are currently very effective in prediction tasks of tabular data \cite{huang2020tabtransformer,grinsztajn2022why}. 
Researchers started attempting to devise neural networks (NN) for modeling tabular data. Early works included FNN and SNN \cite{zhang2016deep}, PNN \cite{qu2016product}, DeepFM \cite{guo2017deepfm} and DNN for YouTube Recommendations \cite{covington2016youtube}. Most of them focused on pre-processing categorical features to adapt to NN architectures, and numerical features were discussed less in these works. TabNN \cite{ke2018tabnn} dealt with both categorical and numerical features to tackle this gap. More recently, some studies introduced the attention mechanism to NN architectures for modeling tabular data to capture information on feature interactions, such as NON \cite{luo2020network}, TabNet \cite{arik2021tabnet}, and SAINT \cite{somepalli2021saint}. 
TabTransformer \cite{huang2020tabtransformer} learned contextual embeddings of categorical features with Transformer~\cite{vaswani2017attention}. 
FT-Transformer \cite{gorishniy2021revisiting} used Transformer to deal with both numerical and categorical features and found a strong effect in a wide range of tabular data. Furthermore, a study found that an appropriate combination of regularization approaches also can make multiple layer perception (MLP) useful for modeling tabular data \cite{kadra2021well}.

\textbf{Graph Neural Networks and Graph Structure Learning.} 
Graph neural networks (GNNs) can produce embeddings for instances by utilizing their own information and recursively aggregating messages from neighbors \cite{gilmer2017neural}. With GNNs, instance associations and feature interactions can be modeled in a more generalized perspective. In this way, GNNs can bring us not only the improvement of prediction performance \cite{Dong2020GSP} but also other achievements, such as handling missing data \cite{You2020GRAPE} and the potential in feature extrapolation \cite{wu2021fate}. Typical GNN methods include 
GCN \cite{kipf2016semi}, GraphSAGE \cite{hamilton2017inductive}, graph attention networks (GAT) \cite{velivckovic2017graph}, and graph isomorphism networks (GIN) \cite{xu2018powerful}.
Although GNNs are strong in learning expressive node embeddings, they are available only if graph-structured data is available \cite{zhu2021gslsurvey, chen2020IDGL}. Besides, the quality of a graph highly correlates with GNN performance where it is used \cite{zhu2021gslsurvey}. A noisy graph structure can harm GNN's performance considerably. Therefore, researchers focused on graph structure learning (GSL) to learn well-structured graphs \cite{zhu2021gslsurvey}. Some metric-based methods adopt kernel functions to calculate the similarity between two nodes as their edge weight. Since kernel functions are differentiable, most metric-based methods can be trained with the end-to-end setting, including AGCN \cite{park2020agcn}, GRCN \cite{yu2020graph}, IDGL \cite{chen2020IDGL}, and SLAPS \cite{Fatemi2021SLAPS}. Some methods directly use the adjacency matrix of the graph as a parameter to be learned, such as LDS \cite{franceschi2019learning}, ProGNN \cite{ProGNN}, and the full graph parameterization learner in SUBLIME \cite{liu2022sublime}.

\textbf{Summary.}
The relevant studies are summarized as Table \ref{tab:related_works}. 
Graph structure learning methods neither model the feature interaction nor deal with pure tabular data. Although non-graph-based methods focus on tabular data learning, they cannot capture the associations among instances. In contrast, our work simultaneously captures feature interactions and models instance associations via graph structure learning for tabular data. 

%% file: 3-Problem.tex
\section{Problem Statement}\label{sec-problem}

Given a tabular dataset depicted as a data matrix $\mathbf{X} \in \mathbb{R}^{n \times m}$ with labels $\mathbf{y} \in \mathbb{R}^n$ but without an available graph structure, where $n$ is the number of data instances, and $m$ is the number of features. Each row $\mathbf{x} \in \mathbf{X}$ is the feature vector of an instance. The goal of this study is to learn an adjacency matrix $\mathbf{A}$, which represents the structure of the graph $\mathcal{G} = (\mathbf{A}, \mathbf{X})$ and indicates associations among instances (i.e., nodes) in the given tabular data, and therefore help improve the performance on the corresponding classification task. In other words, by treating each instance as a node in the graph, we aim to generate the edges between instances that depict their associations. In this way, the original label prediction task can be treated as the node classification task in the learned graph.

%% file: 4-Method.tex
\section{Methodology}\label{sec-method}

\begin{figure*}
	\centering
	\includegraphics[width=1.0\textwidth]{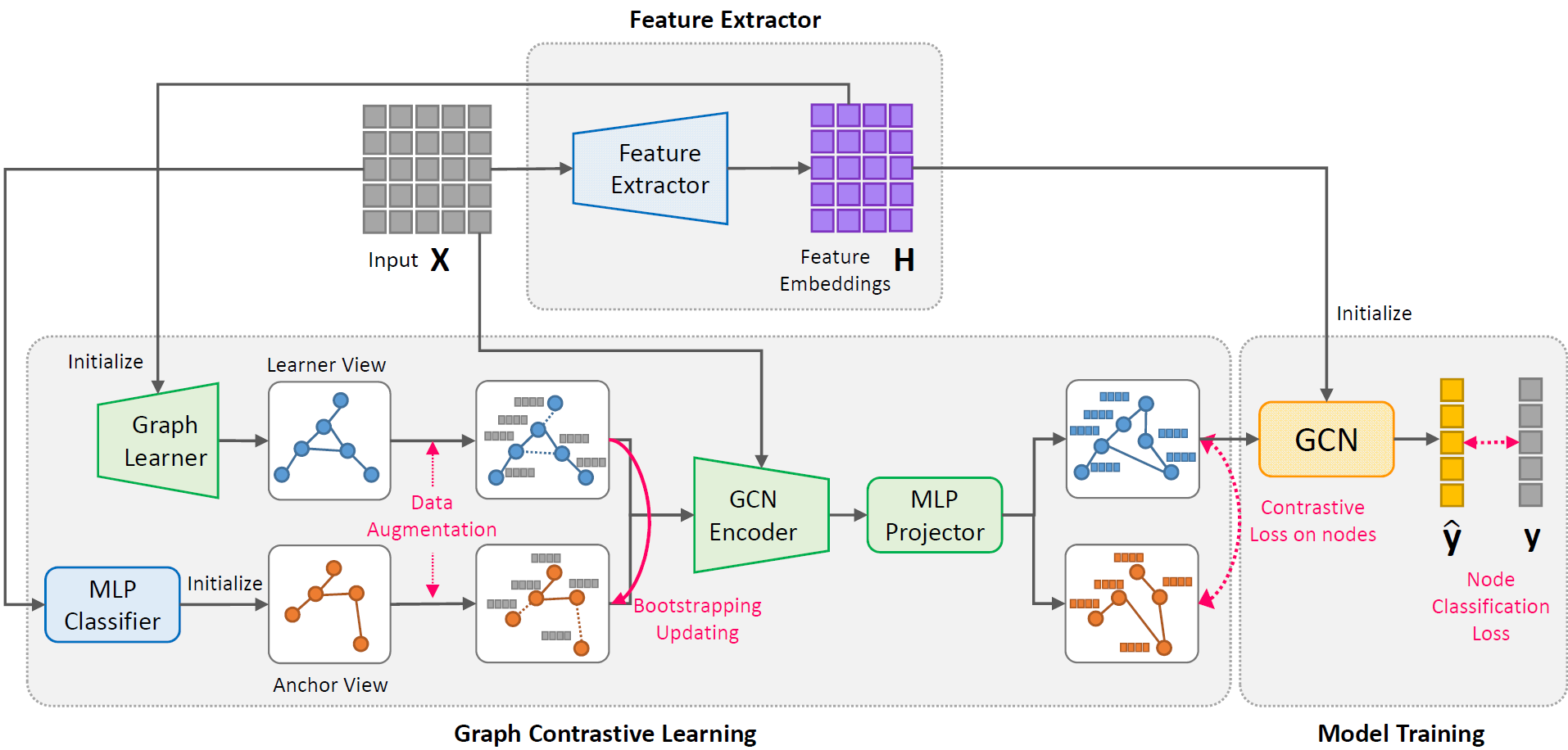}
	\caption{\textbf{The overview of the proposed TabGSL model.} }
	\label{fig:tabgsl_ovw}	
\end{figure*}

The proposed research framework that learns the graph structure for tabular data prediction consists of three main modules: (1) \textit{feature extractor}, (2) \textit{graph structure learning} (GSL), and (3) \textit{model training}. 
The framework overview is presented in Figure~\ref{fig:tabgsl_ovw}.
The feature extractor module, described in Section \ref{FE}, aims to distill useful information and capture the interactions between features. 
The GSL module is composed of the \textit{graph learner} and \textit{graph contrastive learning}, aiming at learning an effective graph structure between instances, which is presented in Section~\ref{sec-GSL}.
The downstream classifier module described in Section~\ref{sec-downcls} is to produce the predictions for the target labels.

\subsection{Feature Extractor}\label{FE}

One possible reason why tree-based models generally outperform deep learning on tabular data is that they can extract useful information from raw features and find the effective interactions of features~\cite{grinsztajn2022why}. The superiority of tree-based models is much clear when most features are categorical.
To encode features appropriately, 
we utilize a feature extractor $\mathcal{FE}(\cdot)$ to produce the feature representation as the transformation of raw features before learning the graph structure. 
$\mathcal{FE}$ consists of the feature tokenizer and $L^{(fe)}$ Transformer layers $Trm^{(l)}(\cdot)$, and eventually generate feature embeddings, $\mathbf{H}$. This can be depicted via $\mathbf{H}^{(l)} = Trm^{(l)}(\mathbf{H}^{(l-1)}; \Theta_{fe}^{(l)})$. The initial hidden states for the first Transformer layer are given by: $\mathbf{H}^{(0)} = stack[[CLS], \tilde{\mathbf{H}}]$, where $[CLS]$ is the classification token. $\tilde{\mathbf{H}}$ is the fusion of initial categorical and numerical feature embeddings, given by $\tilde{\mathbf{H}}=stack[MLP(\mathbf{X}^{cat}; \Theta_{fe}^{cat}), MLP(\mathbf{X}^{num}; \Theta_{fe}^{num})]$, where 
$MLP$ stands for multi-layer perception, $\mathbf{X}^{cat}$ and $\Theta_{fe}^{cat}$ represent the categorical features and the corresponding set of learnable parameters for feature learning via MLP, $\mathbf{X}^{num}$ and $\Theta_{fe}^{num}$ depict the numerical features and the corresponding set of trainable parameters for feature learning via MLP, and $\Theta_{fe}$ is the set of learnable weights of the transformer. 

\subsection{Graph Structure Learning}
\label{sec-GSL}
\textbf{Graph Learner.} We construct the graph learner $\mathcal{GL}(\cdot)$ to find the useful associations between instances based on their features, and eventually generate the weighted adjacency matrix depicting the derived graph.
We adopt MLP to implement $\mathcal{GL}(\cdot)$ to learn the graph structure. Specifically, $\mathcal{GL}(\cdot)$ first receives the derived feature embedding $\mathbf{H}$ from the feature extractor.
Then $\mathcal{GL}(\cdot)$ computes the pairwise similarities between instances through a non-parametric metric function $\phi(\cdot)$, and utilizes the similarity scores as the edge weights of instance pairs in the adjacency matrix $\mathring{\mathbf{A}}$ of the initial graph, given by: $\mathring{\mathbf{A}} = \mathcal{GL}(\mathbf{H}) = \phi(MLP^{L^{(gl)}}(\mathbf{H}))$,
where $L^{(gl)}$ is the number of $\mathcal{GL}(\cdot)$ layers, and $\phi(\cdot)$ is a non-parametric metric function (i.e., cosine similarity function). Here we utilize an MLP as $\mathcal{GL}(\cdot)$ to learn non-linear combinations of feature embeddings to capture the hidden interactions among features. 

\textbf{Contrastive Learning.} We exploit multi-view graph contrastive learning to guide the graph learner and improve the quality of learning latent representations of instances. 
We create two views for the graph learner: the \textbf{anchor view} and the \textbf{learner view}. Two graph views are propagated to generate instance node embeddings and then used to compute the contrastive loss of nodes in graph learning. 
An \textbf{anchor view} is treated as a ``teacher'' to lead the learning of the graph structure stably. Since no graph structure exists originally for tabular data, an initial adjacency matrix should be provided for the anchor view. 
Since the teacher needs to acquire confident label knowledge to better guide graph structure learning, we train a classifier based on original tabular data and utilize the prediction probabilities on labels to initialize the adjacency matrix of anchor view.  
As a larger edge weight between nodes indicates a higher possibility that such two nodes have the same label, node pairs with similar predictions provided by a classifier are supposed to have edges in a graph. 
Specifically, we first train an MLP classifier $Cls(\cdot)$ with original features. Then we determine the entry between nodes $i$ and $j$ in the anchor view's matrix via $\mathbf{A}^{anchor}_{i,j} = \phi [\mathbf{p}_{i}, \mathbf{p}_{j}]$,
where $\mathbf{p}_{i} = Cls(\mathbf{x}_i) \in \mathbb{R}^C$ denotes instance $i$'s predicted probabilities of all $C$ classes by $Cls$, 
and $\phi(\cdot)$ denotes the cosine similarity function. Eventually, we can obtain the anchor view $\mathcal{G}_{anchor} = (\mathbf{A}^{anchor}, \mathbf{X})$ for the follow-up graph learner. 

A \textbf{learner view} is treated as a ``student'' to learn the graph structure. The initial graph of the learner view is established using the previously learned graph $\mathring{\mathbf{A}}$ by $\mathcal{GL}$ with the original feature matrix, given by: 
$\mathcal{G}_{learner} = (\hat{\mathbf{A}}, \mathbf{X})$,
where $\hat{\mathbf{A}} = kNN(\mathring{\mathbf{A}})$. That said, the learner view is initialized by $k$-nearnest neighbors (kNN) based on $\mathring{\mathbf{A}}$, and we keep the edges with top-$k$ connection values.
The learner view's parameters are initialized as an identity matrix or a vector with all elements 1. 
In training, we have two sets of parameters for $\mathcal{GL}(\cdot)$ in the anchor view and learner view, and the learned adjacency matrices are updated by gradient descent. 

To better exploit the learned graph structure, which is supposed to model the correlation between instances, to construct a knowledgeable teacher, we adopt the \textit{structure bootstrapping mechanism}~\cite{caron2018deep, grill2020bootstrap,liu2022sublime} to bring the learned structure into the anchor structure, given by $\mathbf{A}^{anchor} \leftarrow \tau \mathbf{A}^{anchor} + (1 - \tau)\mathbf{A}^{learner}$, 
where $\tau \in [0, 1]$ is a decay rate of the anchor structure. A lower $\tau$ indicates more information about the anchor structure would be changed in training. In this study, we treat $\tau$ as a tuning hyperparameter.
Adding information gradually about the learned structure into the anchor structure can be viewed as data augmentation.

Two augmented graph views are propagated to a Graph Convolutional Network (GCN)~\cite{kipf2016semi} encoder and an MLP projector to generate node embeddings for contrastive learning. 
We make projected node embeddings from two views similar by maximizing their contrastive loss. The symmetric normalized temperature-scaled cross-entropy (NT-Xent)~\cite{oord2018representation, sohn2016improved} is used as the contrastive loss, given by: $\mathcal{L}_{gcl} = \frac{1}{2n}\sum_{i=1}^n [l(\mathbf{z}_{i}^{anchor}, \mathbf{z}^{learner}_{i}) + l(\mathbf{z}^{learner}_{i}, \mathbf{z}^{anchor}_{i})]$, 
where $l(\mathbf{z}^{learner}_{i}, \mathbf{z}^{anchor}_{i}) = \log\frac{\exp[\ \phi(\mathbf{z}^{learner}_{i}, \mathbf{z}^{anchor}_{i})/t]}{\sum_{j \neq i}\exp[\ \phi(\mathbf{z}^{learner}_{i}, \mathbf{z}^{anchor}_{j})/t]}$,
where 
$\mathbf{z}^{anchor}_{i}$ and $\mathbf{z}^{learner}_{i}$ are node $i$'s embeddings obtained from the anchor and the learner views, 
and $t$ is the temperature hyper-parameter.

\textbf{Data Augmentation.}
To bring robust model training and to better learn the association between instances, we inject two data augmentation mechanisms into graph structure learning: feature masking and edge dropping. 
For feature masking, a subset of features is selected randomly and masked before generating node embeddings via GSL. This is depicted by $\mathbf{X'}^{anchor} = [\mathbf{x'}_1 \odot \mathbf{m}^{anchor}, ..., \mathbf{x'}_{n} \odot \mathbf{m}^{anchor}]^{\top}$ and $\mathbf{X'}^{learner} = [\mathbf{x'}_1 \odot \mathbf{m}^{leaner}, ..., \mathbf{x'}_{n} \odot \mathbf{m}^{leaner}]^{\top}$ for anchor and learner views, respectively, where
$\mathbf{m}^{anchor}$ and $\mathbf{m}^{learner}$ are mask vectors for two views sampled from Binominal distributions with ratio hyper-parameters $\rho^{anchor}$, $\rho^{learner} \in [0, 1]$ that determine how many features to mask for anchor and learner, and $\mathbf{x'}_i$ is the $i$-th row feature vector. 
the graph structure is corrupted by randomly dropping a subset of edges: $\mathbf{A'} = \mathbf{A} \odot \mathbf{M}^{(e)}$, where $\mathbf{M}^{(e)}$ is a mask matrix with elements $\mathbf{M}_{i, j} \sim \mathcal{B}er(\rho_{edge})$, $\mathcal{B}er(\rho_{edge})$ stands for the Bernoulli distribution with the probability $\rho_{edge}$, and $\odot$ is the Hadamard operation. Both anchor and learner views are processed by the augmentations of feature masking and edge dropping.

\subsection{Modeling Training}
\label{sec-downcls}
To produce the prediction outcomes, we feed the adjacency matrix $\dot{A}$ of the learned graph derived from the learner view into a $L^{(mt)}$-layer GCN for training the node classification model. The learned embeddings from the feature extractor, $\mathbf{H}$, are used as the initial node vectors. 
Together with an MLP to generate final representations of instances, the prediction probability of instance $i$ on class label $c$ can be depicted via: $\hat{y}_{i,c}=softmax(ReLU[\bar{\mathbf{D}}^{\frac{-1}{2}} \bar{\mathbf{A}} \bar{\mathbf{D}}^{\frac{-1}{2}} \bar{\mathbf{Z}}^{(l-1)} \Theta_{mt}^{(l)}])$, where $\bar{\mathbf{Z}}^{(l)}$ is the $l$-th layer node emebdding matrix, $\bar{\mathbf{Z}}^{(0)}=\mathbf{H}$, $\bar{\mathbf{A}}$ is $\mathbf{A}$ with self-loops, and $\bar{\mathbf{D}}$ is the degree matrix of $\mathbf{A}$.

We construct the model training in an end-to-end manner. The training objective consists of graph contrastive learning $\mathcal{L}_{gcl}$ and node classification $\mathcal{L}_{nc}$, depicting as minimizing the loss function: $\mathcal{L}=\mathcal{L}_{nc}-\mathcal{L}_{gcl}$.
We take the negative value of $\mathcal{L}_{gcl}$ owing to enforcing embeddings of two views to get closer, i.e., maximizing the contrastive loss.
We train node classification by minimizing the negative likelihood loss, i.e., $\mathcal{L}_{nc} = - \frac{1}{n} \sum_{i=1}^{n} \sum_{c=1}^{C} y_{i, c}$, where $y_{i, c} \in \{0, 1\}$ indicates if the ground-truth label of instance $i$ is $c$.
The training objective is optimized by the adaptive moment estimation (Adam)~\cite{kingma2014adam}.

%% file: 5-Experiments.tex
\section{Experiments}
\label{sec-exp}

\subsection{Evaluation Setup}
\textbf{Datasets and Metric.}
We use datasets from the OpenML-CC18 benchmark~\cite{openml}, which contains thousands of real-world classification datasets manually curated for benchmarking. We select $30$ datasets that contain both numerical and categorical features. The statistics of datasets are displayed in Table~\ref{tab:data} in Appendix. Each dataset is split into training, validation, and test sets with the ratio $70\%$, $15\%$, and $15\%$. Models are trained on the training sets for $5$ trials with different random seeds. The validation sets are used for tuning hyper-parameters. The test sets are used to evaluate models with the tuned hyper-parameters, in which average scores of model performances are reported. 
Since selected datasets are not balanced, we use the F1-score of the minority label class as the evaluation metric for the datasets with binary labels. For multi-class datasets, the macro F1-score is used.

\textbf{Baselines.}
We include two categories of models as baselines in this study, (1) tree-based models and (2) deep neural networks (DNN) for tabular data. There are three typical tree-based models and four well-known DNN-based methods for tabular data involved, namely LightGBM \cite{lightgbm}, XGBoost \cite{xgboost}, CatBoost \cite{catboost}, MLP, FT-Transformer (FT-T) \cite{gorishniy2021revisiting}, SubTab \cite{ucar2021subtab}, and SUBLIME~\cite{liu2022sublime}. For the MLP baseline, we adopt the architecture used in \cite{gorishniy2021revisiting}, which is well-designed and found to be a strong baseline. 
SubTab \cite{ucar2021subtab} is an unsupervised method to learn representations from tabular data with ensemble learning by dividing features into subsets, and takes logistic regression (for binary classification) and MLP (for multi-class classification) to be the downstream classifier. 
FT-Transformer (FT-T) \cite{gorishniy2021revisiting} is a simple adaptation of the Transformer architecture that outperforms other DL solutions on tabular datasets.
SUBLIME~\cite{liu2022sublime} is an unsupervised graph structure learning method that can accept tabular data as model input. With the learned instance embeddings, we utilize MLP as the downstream classifier.
For the hyperparameters of all baseline methods, we refer to their original papers and follow their tuning strategies for choosing the best performance.

\textbf{Computing Settings.}
All methods were conducted with Python (version 3.7) on a Linux Intel Xenon Gold 6138 processor with 2.0 GHz RAM and a 450G CPU. For neural network-based methods, an NVIDIA Tesla V100 32 GB GPU was used for speeding up the training and the PyTorch package \cite{paszke2019pytorch} was used to implement. 

    \begin{table}[!t]
      \caption{\textbf{Tuning Ranges of Hyper-parameters.} 
      \textit{$\mathcal{FE}$ = feature extractor; GSL = graph structure learning; NC = node classification.}}
      \centering
      \begin{tabular}{l | c c }
        \toprule
         \textbf{Hyper-parameter} & \textbf{Category} & \textbf{Tuning Range} \\
        \hline
        
        $d_{fe}$ (embedding dimension) & $\mathcal{FE}$ & Integers in [16, 32, 64, 128, 256, 512] \\
        $L^{(fe)}$ & $\mathcal{FE}$ & [1, 2, 3, 4] \\
        
        \hline
        $k$ & GSL & [5, 10, 15, 20, 25, 30, 35] \\
        $d_{gl}$ (embedding dimension) & GSL & [64, 128, 256] \\
        $L^{(gl)}$ & GSL & 2, 3 \\
        $\tau$ & GSL & [.99, .999, .9999, .99999, 1] \\
        $\rho_{anchor}$ & GSL & Real numbers in [0.6, 0.75) \\
        $\rho_{learner}$ & GSL & Real numbers in [0, 0.7) \\
        $\rho_{edge}$ & GSL & Real numbers in [0.25, 0.55) \\
        
        \hline
        
        Learning rate & Train & Real numbers in [5e-4, 5e-3) \\
        Weight decay rate & Train & Real numbers in [0, 1e-5) \\
        Dropout rate & Train & Real numbers in [0.4, 0.8) \\
        $t$ & Train & 0.2, 0.3, 0.4 \\
        
        \hline
        
        $L^{(mt)}$ & NC & 2, 3 \\
        $d_{mt}$ (embedding dimension) & NC & [16, 32, 64, 128] \\
        Learning rate & NC & Real numbers in [5e-4, 5e-3) \\
        Weight decay rate & NC & Real numbers in [0, 1e-5) \\
        Dropout rate & NC & Real numbers in [0.4, 0.6, 0.8) \\
        
        \bottomrule
    \end{tabular}\label{tab:HP}
    \end{table}

\textbf{Hyper-parameters Tuning.}
By following the construction of training cocktails for tabular data~\cite{kadra2021well}, we tune hyper-parameters with multi-fidelity Bayesian optimization method (BOHB), which combines Hyperband \cite{li2017hyperband} and Bayesian Optimization \cite{mockus1994application, BOHB} due to its low time cost, widely strong performance, and ability to deal with the categorical and conditional hyper-parameters. We use Optuna \cite{optuna} to implement BOHB. Different sets of hyper-parameters are tubed for each dataset for different training settings. Hyper-parameters for GSL methods that we tune and their tuning ranges are presented in Table~\ref{tab:HP}. Hyper-parameters of baselines are also tuned by Optuna in the ranges suggested by existing studies that focused on comparisons of tree-based and neural network-based methods for tabular data \cite{gorishniy2021revisiting,ucar2021subtab,lightgbm, xgboost, catboost}.

\subsection{Experiment Results}

    \begin{table*}[!t]
      \caption{\textbf{Performance Comparison in F1-score.} F1-scores or macro F1-scores are reported as mean±std (\%). The best method's score of each dataset is highlighted in bold and red. The standard deviations are not considered for determining the highlighted scores. The second-best method is highlighted in bold.} 
      \centering
      \resizebox{\textwidth}{!}{%
  \begin{tabular}{l | r r r r r r r r}
    \toprule
    Data & LightGBM & XGBoost & CatBoost & MLP & FT-T & SubTab & SUBLIME & TabGSL \\
    \hline
    23 & 48±0.0 & 48±1.4 & 47±1.1 & 37±4.2 & \textbf{49±2.1} & 24±4.9 & 45±3.0 & \textcolor{red}{\textbf{52±0.2}} \\
    
    31 & 43±7.4 & 51±4.7 & 32±8.7 & 57±4.4 & \textbf{58±2.4} & 8±10.2 & 55±2.3 & \textcolor{red}{\textbf{59±2.0}} \\
    
    48 & 17±0.0 & \textbf{51±6.0} & 43±4.5 & 36±5.3 & 42±5.2 & 23±16.5 & 51±6.6 & \textcolor{red}{\textbf{56±2.8}} \\
    
    446 & 82±1.3 & 88±6.2 & 88±1.9 & \textbf{97±2.2} & 96±7.0 & 60±13.4 & \textcolor{red}{\textbf{100±0.0}} & \textcolor{red}{\textbf{100±0.0}} \\
    
    475 & 26±3.0 & 23±6.6 & 28±2.4 & 29±7.0 & \textbf{34±5.8} & 20±4.8 & 16±3.2 & \textcolor{red}{\textbf{42±2.9}} \\
    
    720 & \textcolor{red}{\textbf{80±0.4}} & \textcolor{red}{\textbf{80±0.3}} & \textbf{75±0.1} & \textcolor{red}{\textbf{80±0.3}} & \textcolor{red}{\textbf{80±0.5}} & 55±27.3 & \textcolor{red}{\textbf{80±0.3}} & \textcolor{red}{\textbf{80±0.4}} \\
    
    825 & 85±2.5 & 84±2.3 & 83±2.3 & \textbf{87±1.8} & 85±2.3 & 78±4.8 & \textcolor{red}{\textbf{88±1.4}} & \textcolor{red}{\textbf{88±0.5}} \\
    
    853 & 78±5.1 & 80±3.5 & \textcolor{red}{\textbf{86±1.2}} & \textbf{85±1.5} & 83±4.3 & 57±32.7 & \textcolor{red}{\textbf{86±1.9}} & \textcolor{red}{\textbf{86±3.0}} \\
    
    902 & 66±4.7 & 56±8.4 & 74±7.4 & 67±8.2 & \textbf{80±3.5} & 65±2.2 & 62±2.5 & \textcolor{red}{\textbf{84±0.0}} \\

    915 & 14±7.7 & 41±5.7 & 15±12.8 & \textbf{45±3.8} & \textbf{45±13.7} & 26±16.1 & 41±8.3 & \textcolor{red}{\textbf{57±2.2}} \\

    941 & 53±3.8 & 70±3.8 & 70±3.6 & \textbf{73±3.4} & 72±2.2 & 45±17.6 & 68±4.5 & \textcolor{red}{\textbf{78±2.6}} \\
    
    955 & 35±13.8 & 52±10.0 & 67±5.5 & \textbf{74±5.2} & 70±11.3 & 19±3.2 & 62±15.6 & \textcolor{red}{\textbf{77±4.9}} \\
    
    983 & \textbf{61±1.1} & 59±2.5 & 57±2.1 & 56±4.4 & \textcolor{red}{\textbf{64±0.8}} & 10±17.4 & 60±2.0 & \textcolor{red}{\textbf{64±3.5}} \\
    
    1006 & 72±5.8 & \textbf{74±3.9} & 65±3.3 & 67±6.0 & 72±4.0 & 58±10.3 & 67±9.0 & \textcolor{red}{\textbf{78±5.0}} \\
    
    1012 & 43±4.4 & 56±5.2 & 58±9.6 & 72±3.8 & 66±10.5 & 29±15.3 & \textbf{73±1.5} & \textcolor{red}{\textbf{76±3.3}} \\
    
    1115 & 17±0.0 & \textbf{68±4.1} & 60±4.8 & 52±8.3 & 54±5.3 & 34±8.7 & 60±4.7 & \textcolor{red}{\textbf{74±2.8}} \\
    
    1167 & 6±6.0 & 13±10.3 & 4±10.3 & 17±7.0 & 21±7.8 & 13±4.5 & \textbf{24±5.5} & \textcolor{red}{\textbf{33±5.8}} \\
    
    1498 & 53±3.8 & 40±6.2 & 48±5.2 & 58±4.5 & \textbf{59±4.2} & 19±14.1 & 51±3.8 & \textcolor{red}{\textbf{64±5.8}} \\
    
    1549 & 10±1.6 & \textbf{19±3.0} & 10±1.8 & 7±2.8 & 14±2.4 & 7±0.9 & 12±3.4 & \textcolor{red}{\textbf{20±5.1}} \\

    1552 & 34±3.0 & \textbf{38±1.8} & 32±3.3 & 30±2.8 & 36±4.4 & 20±8.8 & 34±1.9 & \textcolor{red}{\textbf{44±1.7}} \\
        
    1553 & 45±0.0 & 46±5.0 & \textbf{47±4.7} & 39±4.4 & 42±2.1 & 30±6.9 & 40±1.9 & \textcolor{red}{\textbf{55±1.8}} \\
        
    1555 & 13±2.7 & \textcolor{red}{\textbf{19±3.8}} & 15±2.0 & 11±2.9 & \textbf{17±1.6} & 4±0.1 & 16±1.7 & \textcolor{red}{\textbf{19±1.9}} \\
        
    1557 & 65±1.1 & \textbf{65±0.7} & 65±1.1 & 64±1.1 & 64±1.5 & 40±11.9 & 62±0.3 & \textcolor{red}{\textbf{66±0.2}} \\
        
    40663 & 37±1.8 & \textbf{55±3.2} & 46±3.7 & 53±6.4 & 55±3.7 & 11±12.4 & 54±4.8 & \textcolor{red}{\textbf{59±1.8}} \\
        
    40705 & \textbf{87±0.3} & \textbf{87±1.0} & \textcolor{red}{\textbf{88±1.5}} & 86±2.2 & 86±2.5 & 83±1.1 & 83±1.8 & \textcolor{red}{\textbf{88±0.0}} \\
        
    40710 & \textbf{85±1.5} & 84±4.9 & \textbf{85±3.5} & 84±2.4 & 82±2.0 & 64±3.1 & 80±1.3 & \textcolor{red}{\textbf{88±0.0}} \\
        
    40981 & \textbf{84±1.4} & 82±1.9 & 82±0.4 & 50±43.8 & 81±2.1 & 78±11.0 & \textcolor{red}{\textbf{85±2.9}} & \textcolor{red}{\textbf{85±0.0}} \\
        
    43255 & 82±1.9 & 81±1.2 & 81±1.9 & \textbf{87±1.1} & 85±1.7 & 65±0.4 & 86±2.4 & \textcolor{red}{\textbf{88±1.5}} \\
        
    43942 & \textcolor{red}{\textbf{100±0.0}} & 99±0.7 & \textcolor{red}{\textbf{100±0.0}} & 96±3.4 & 99±0.5 & 67±5.1 & 85±1.2 & \textbf{98±0.6} \\
        
    44098 & 47±9.0 & 57±5.3 & 25±8.8 & \textbf{62±3.1} & \textbf{62±3.5} & 7±7.4 & 59±1.2 & \textcolor{red}{\textbf{65±1.1}} \\
        \bottomrule
    \end{tabular}}
    \label{tab:main1}
    \end{table*}

\begin{figure}
    	\centering
    	\includegraphics[width=0.8\textwidth]{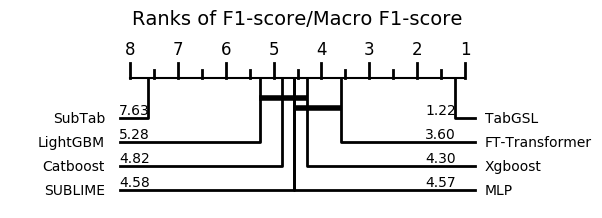}
    	\caption{\textbf{Critical Difference Diagrams of The Wilcoxon Signed-Rank Test for \textit{TabGSL} and All Baselines.}  
     The difference between the two methods is not significant if their index lines are connected or passed through by a bold bar.}
    	\label{fig:CD}	
    \end{figure}

The results of the performance comparison are presented in Table~\ref{tab:main1}. In addition to reporting the detailed scores, we utilize the Critical Difference (CD) diagram of the ranks using the F1-score or macro F1-score with the Wilcoxon signed-rank test \cite{wilcoxon}, which is commonly used to compare model performances among a large number of datasets in existing studies \cite{kadra2021well, demvsar2006statistical}. The significant levels of tests are set as $0.05$. The plots are displayed in Figure~\ref{fig:CD}.

\textit{Evaluation Question 1: Can learning graph structure from tabular data improve classification performance? Do graph neural network-based methods outperform tree-based models?}

Based on the results shown in Table~\ref{tab:main1} and Figure~\ref{fig:CD}, we can see that the proposed TabGSL consistently outperforms all competing methods, including tree-based models and deep neural network-based methods across almost all datasets. Such results verify the effectiveness of learning the association between instances and modeling feature interactions in the proposed graph structure learning framework. While a recent study had pointed out that irrelevant features can significantly degrade the performance of neural network models~\cite{grinsztajn2022why}, the GNN-based feature aggregation based on learned graph structure can to some degree mitigate such a negative effect, and thus bring performance improvement. The learned graph can help adjust the importance of some features. The learner can properly add edges to instance nodes with similar useful features to strengthen their contribution and disconnect nodes where some of their feature dimensions can hurt performance. In short, such an outcome conveys a key insight: neural network-based methods for tabular data prediction can be improved by learning graph structure among instances.

\begin{figure*}
	\centering
	\includegraphics[width=1\textwidth]{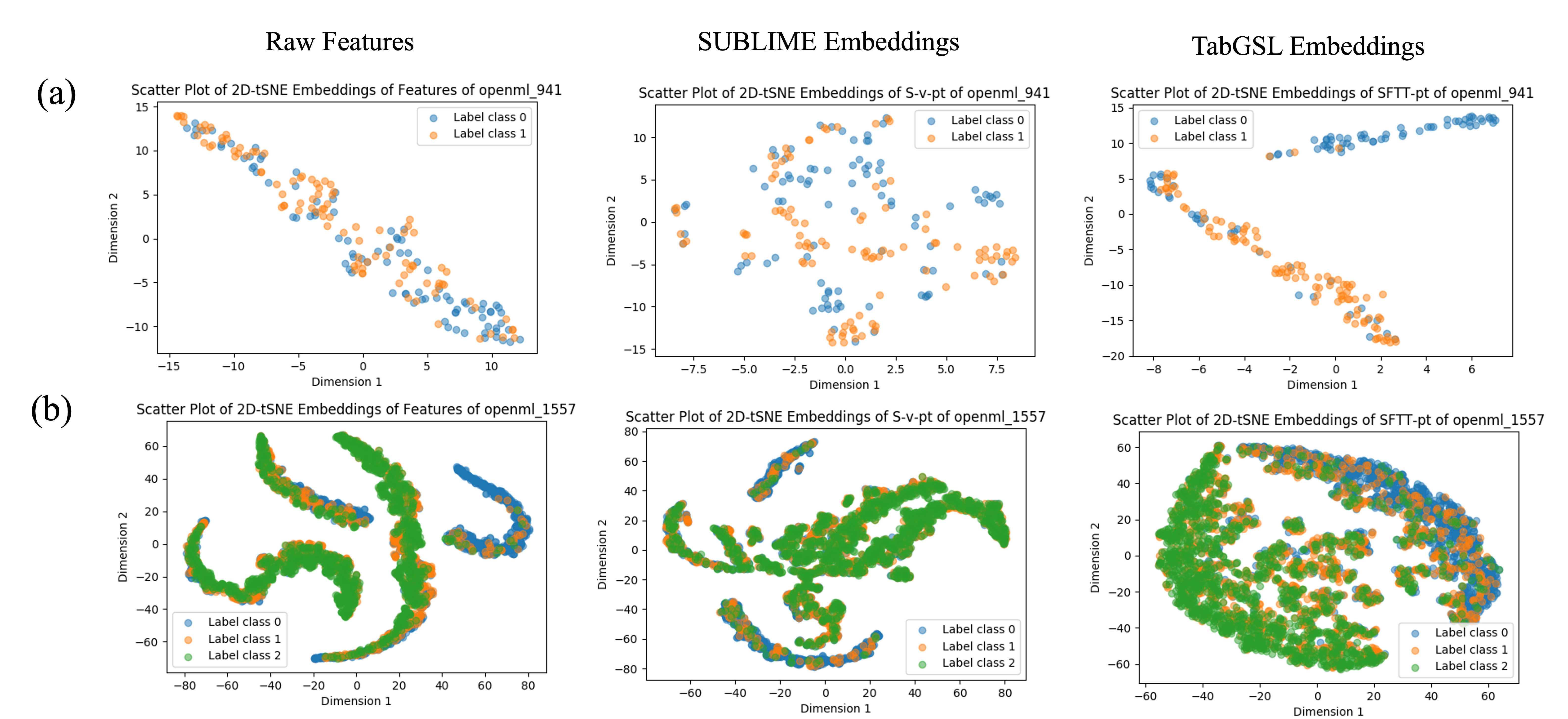}
	\caption{\textbf{Scatter Plots of t-SNE of Raw Features, SUBLIME Embeddings, and TabGSL Embeddings for Datasets 941 and 1557.} \textbf{(a)} Dataset 941. \textbf{(b)} Dataset 1557. 
 }
	\label{fig:tsne}	
\end{figure*}

\textit{Evaluation Question 2: Can the embeddings produced by the proposed TabGSL better separate instances of different classes in the feature space?}

To intuitively demonstrate the effect of graph structure learning for tabular data, we aim at visualizing and comparing raw features, and instance node embeddings generated by GSL including SUBLIME~\cite{liu2022sublime} and the proposed TabGSL. 
We utilize 
t-SNE~\cite{van2008visualizing} to plot the visualization of the feature space, in which each point is a data instance.
Data points with the same colors represent instances with the same class label. In this way, we can observe whether a feature vector or a node embedding captures the knowledge about classification. 
We select two datasets for t-SNE visualization. The results are shown in Figure~\ref{fig:tsne}. 
Generally speaking, points with different colors disperse more obviously for the plots of GSL methods than raw features.
For example, from the three subplots of dataset 941, blue and orange points are mixed together using raw features while they are separated using GSL node embeddings. Furthermore, points with the same color tend to be much closer to each other for node embeddings produced by our TabGSL, compared to SUBLIME.
This finding again supports that learning an effective graph structure can be beneficial for encoding tabular data into embeddings with label knowledge. A similar result can be obtained in 
in multi-class datasets such as dataset 1557. 

%% file: 6-Conclusions.tex
\section{Conclusions}
\label{sec-conclude}

In this work, we find that learning graphs from tabular data is helpful for classification tasks and can make NN models as competitive as tree-based models or even more powerful. The finding is established on the proposed Tabular Graph Structure Learning (TabGSL) method. TabGSL's performance on prediction tasks using GSL, along with the integration with the feature extractor and graph neural network, can improve greatly with the end-to-end training manner. The key is that by capturing contextual information of feature interactions as better initialized embeddings, TabGSL is able to produce the graph structure that better depicts the association between instances. 

\textbf{Limitations.} There are several limitations to this study. First, tabular datasets used in this study are not large, with instances fewer than $5000$ and features fewer than $50$. There may be potential time and space cost problems when applying this study to large-scale datasets. Second, robustness studies can be conducted by identifying the effects of feature masking ratios of anchor and learner views (i.e., $\rho_{anchor}$ and $\rho_{learner}$). In addition, it would be complete if labels were contaminated to test the robustness of GSL methods. We remain these limitations to be solved as future work.

%% file: A-Appendix.tex
\newpage
\appendix

\section*{Appendix}

\section{Data Statistics}
\label{apd-data}
We present the statistics of the $30$ OpenML-CC18 benchmark datasets~~\cite{openml} used for running the experiments in Table \ref{tab:data}. 

    \begin{table*}[h]
      \caption{\textbf{Statistics of Datasets Used in This Study.} $\#$ = Number of; Num. = Numeric; Cat. = Categorical; Maj. = Majority; Min. = Minority.}
      \centering
      \begin{tabular}{r | c c c c c c c}
        \toprule
          &  &  & \textbf{$\#$Num.} & \textbf{$\#$Cat.} &  & \textbf{Maj. Class} & \textbf{Min. Class} \\
          \textbf{ID} & \textbf{$\#$Instance} & \textbf{$\#$Feature} & \textbf{Feature} & \textbf{Feature} & \textbf{\#Class} & \textbf{Size} & \textbf{Size} \\
        \hline
        
        \textbf{23} & 1473 & 9 & 2 & 7 & 3 & 629 & 333 \\
        \textbf{31} & 1000  & 20  & 7  & 13  & 2  & 700  & 300 \\
        \textbf{48} & 151  & 5  & 3  & 2  & 3  & 52  & 49 \\
        \textbf{446} & 200  & 7  & 6  & 1  & 2  & 100  & 100 \\
        \textbf{475} & 400  & 5  & 1  & 4  & 4  & 100  & 100 \\
        \textbf{720} & 4177  & 8  & 7  & 1  & 2  & 2096  & 2081 \\
        \textbf{825} & 506  & 20  & 17  & 3  & 2  & 283  & 223 \\
        \textbf{853} & 506  & 13  & 12  & 1  & 2  & 297  & 209 \\
        \textbf{902} & 147  & 6  & 2  & 4  & 2  & 78  & 69 \\
        \textbf{915} & 315  & 13  & 10  & 3  & 2  & 182  & 133 \\
        \textbf{941} & 189  & 9  & 2  & 7  & 2  & 99  & 90 \\
        \textbf{955} & 151  & 5  & 3  & 2  & 2  & 99  & 52 \\
        \textbf{983} & 1473  & 9  & 2  & 7  & 2  & 844  & 629 \\
        \textbf{1006} & 148  & 18  & 3  & 15  & 2  & 81  & 67 \\
        \textbf{1012} & 194  & 28  & 2  & 26  & 2  & 125  & 69 \\
        \textbf{1115} & 151  & 6  & 2  & 4  & 3  & 52  & 49 \\
        \textbf{1167} & 320  & 8  & 7  & 1  & 2  & 213  & 107 \\
        \textbf{1498} & 462  & 9  & 8  & 1  & 2  & 302  & 160 \\
        \textbf{1549} & 750  & 40  & 37  & 3  & 8  & 165  & 57 \\
        \textbf{1552} & 1100  & 12  & 8  & 4  & 5  & 305  & 153 \\
        \textbf{1553} & 700  & 12  & 8  & 4  & 3  & 245  & 214 \\
        \textbf{1555} & 1000  & 40  & 37  & 3  & 8  & 240  & 89 \\
        \textbf{1557} & 4177  & 8  & 7  & 1  & 3  & 1447  & 1323 \\
        \textbf{40663} & 399  & 32  & 12  & 20  & 5  & 96  & 44 \\
        \textbf{40705} & 959  & 44  & 42  & 2  & 2  & 613  & 346 \\
        \textbf{40710} & 303  & 13  & 5  & 8  & 2  & 165  & 138 \\
        \textbf{40981} & 690  & 14  & 6  & 8  & 2  & 383  & 307 \\
        \textbf{43255} & 1000  & 7  & 3  & 4  & 2  & 518  & 482 \\
        \textbf{43942} & 898  & 38  & 6  & 32  & 2  & 486  & 412 \\
        \textbf{44098} & 1000  & 20  & 7  & 13  & 2  & 700  & 300 \\
        \bottomrule
    \end{tabular}\label{tab:data}
    \end{table*}


\section{Analysis of Training Strategies}
\label{apd-train}
When constructing neural network-based models for tabular data, existing studies have shown various but effective training strategies. Self-supervised pre-training and supervised fine-tuning in VIME~\cite{vime20icml}, TabNet~\cite{arik2021tabnet}, and SCARF~\cite{bahri2022scarf}. Two-stage training, training a self-supervised model to obtain embeddings and building a downstream classifier using embeddings, is verified to be very effective in SubTab~\cite{ucar2021subtab} and SUBLINE~\cite{liu2022sublime}. Besides, incorporating self-supervised learning with the supervised signal as an end-to-end training manner also brings performance improvement on tabular data~\cite{kossen2021self,Rubachev2022revisiting}. Therefore, we wonder which kind of training strategy is more proper and effective for the proposed TabGSL.

The default training strategy of the proposed TabGSL model is \textit{end-to-end}. We investigate and compare two more training strategies. One is \textit{Two-stage Training} (Two-Stage), and the other is \textit{Pre-Training \& Fine-Tune} (PT-FT). For the strategy of two-stage training, the goal is to make the model concentrate on learning a better graph structure among instances. We first train the first two modules, including feature extractor and graph structure learning, to produce effective feature representations for instances and derive the learned graph structure that depicts the latent correlation between instances. Then the obtained embeddings $\mathbf{H}$ from the feature extractor and the adjacency matrix $\dot{A}$ of the learned graph are fed into the final GCN module to train the predictor of class labels. The first stage is only to optimize the graph contrastive learning loss $-\mathcal{L}_{gcl}$, while the second stage is to minimize the classification loss $\mathcal{L}_{nc}$. For the strategy of pre-training \& fine-tuning, the goal is to adjust the learned graph structure based on the downstream task of node classification. We first train the first two modules by optimizing $-\mathcal{L}_{gcl}$, and obtain the corresponding model weights. Such model parameters are further fine-tuned, together with the weight optimization of the last module (i.e., the last GCN component and the MLP projector), through the classification loss $\mathcal{L}_{nc}$.

\begin{figure}[!t]
    	\centering
    	\includegraphics[width=1.0\textwidth]{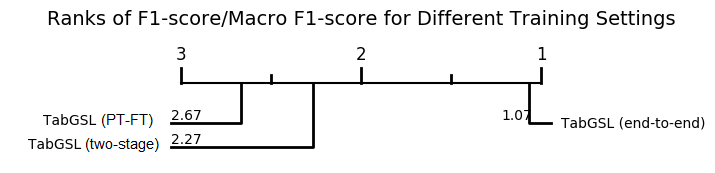}
    	\caption{\textbf{Critical Difference Diagrams of The Wilcoxon Signed-Rank Test for \textit{TabGSL} and its training variants.} 
     The difference between the two methods is not significant if their index lines are connected or passed through by a bold bar.}
    	\label{fig:CD_train}	
    \end{figure}

The results of analysis on various TabGSL training settings are presented in the CD diagram in Figure~\ref{fig:CD_train}. It can be obviously found that the end-to-end training strategy leads to the best performance for TabGSL. We discuss the potential reasons as follows. (a) \textit{Joint Optimization}: In end-to-end training, the graph structure learning and the instance classification are optimized together. This means the graph structure can be directly influenced by the classification task, and the classifier can adapt to changes in the graph structure during training. This joint optimization allows the model to learn a graph structure that is more suited to the specific task. (b) \textit{Consistent Objective}: End-to-end training allows the model to focus on a single, consistent objective function during training. In contrast, the other two methods involve a change in objectives during the process -- from graph structure learning to classification or fine-tuning. This change can sometimes lead to suboptimal performance as the objectives might not be perfectly aligned. 
(c) \textit{Avoid Overfitting}: In the two-stage training and pre-training \& fine-tuning strategies, there is a risk that the model might overfit to the graph structure learning stage, resulting in a model that does not generalize well to the classification task. End-to-end training mitigates this risk as it balances the two tasks from the beginning.

\section{Hyperparameter Sensitivity}
\label{apd-hyperp}
We study how key hyperparameters of TabGSL affect the prediction performance. The hyperparameters we would like to examine include $\tau$ and $k$. 
We select such two hyperparameters because they directly influence the quality of capturing the associations between data instances. In addition, the main goal of this work is to explore the potential ability to learn graph structure from tabular data. Factors that can shape the learned graph are what we need to look into.

\begin{figure}[!t]
    	\centering
    	\includegraphics[width=0.8\textwidth]{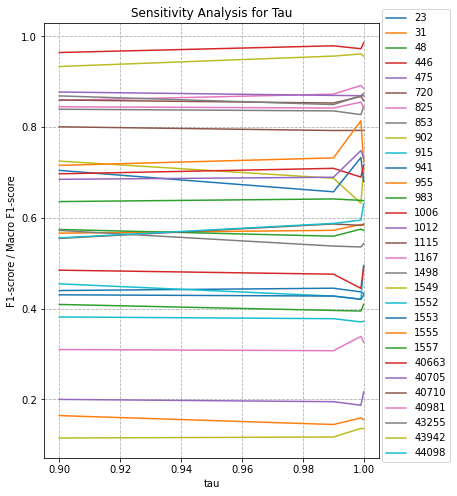}
    	\caption{\textbf{Hyperparameter Analysis on $\tau$} with range: $0.90$, $0.92$, ..., and $1.0$.}
    	\label{fig:hyperp_tau}	
    \end{figure}

\textbf{Structure Bootstrapping Factor $\tau$.} The hyperparameter $\tau$ is used in the implementation of the structure bootstrapping mechanism, i.e., $\mathbf{A}^{anchor} \leftarrow \tau \mathbf{A}^{anchor} + (1 - \tau)\mathbf{A}^{learner}$, where $\tau \in [0, 1]$. It takes charge of bringing the learned structure to update the anchor structure. 
A higher $\tau$ indicates the anchor structure is updated slowly since less information is transferred from the learned structure. We assume that the teacher's knowledge in the anchor view is clean, and the learner view's information at the early learning stage is unstable. Hence, we choose to set $\tau$ as larger values. By varying the $\tau$ value as $0.90$, $0.92$, ..., and $1.0$, we report the performance scores of all datasets in Figure~\ref{fig:hyperp_tau}. We can find that the performance in terms of the F1 score keeps nearly unchanged until $\tau$ is very close to $1.0$, where the performance trends become inconsistent across datasets. Such results bring two insights. First, to have stable training in TabGSL, learning the tabular graph structure requires gradual knowledge updating from the learned topology. The learned structure does carry the latent associations among instances, which benefit the model's generalization ability. Second, the requirement of guidance from the learned graph varies among datasets. Some datasets do not need to have guidance via structure bootstrapping when discovering the effective graph structure increasingly. There are two possible reasons behind this finding. The first is \textit{Simplicity of Structure}. If the inherent structure among data instances is simple and can be easily captured by the learning model, there might be no need for structure bootstrapping. A complex structure bootstrapping mechanism might be overkill and even detrimental for datasets with simple patterns. The second is \textit{Well-structured Datasets}. Some datasets might already be well-structured or contain useful features, meaning that the important features and the correlation between features and class labels are already prominent. In such cases, structure bootstrapping may not offer significant additional benefits. We believe that exploring which kinds of datasets require graph structure learning to what extent will be the key future task of graph structure learning for tabular data.

\begin{figure}[!t]
    	\centering
    	\includegraphics[width=0.8\textwidth]{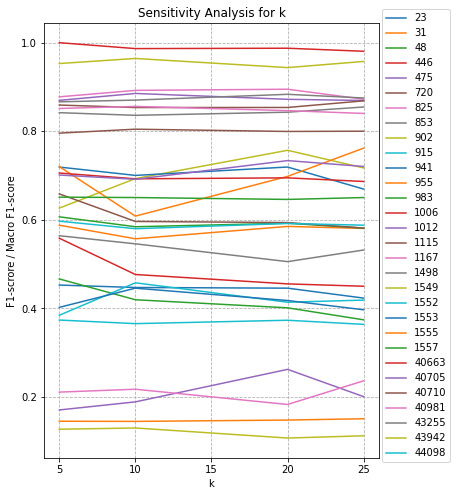}
    	\caption{\textbf{Hyperparameter Analysis on $k$} with values: $5$, $10$, ..., and $25$.}
    	\label{fig:hyperp_k}	
    \end{figure}

\textbf{Initial $k$NN Graph on Learner View.}
In graph contrastive learning, we initialize the learner view's graph using $k$-nearest neighbors ($k$NN) based on the adjacency matrix $\mathring{\mathbf{A}}$ derived from the first graph learner. $k$NN leverages the concept of local similarity, assuming that data points in the feature space will likely share similar characteristics or classes. This can be a beneficial starting point for graph structure learning, as it initiates the graph with a basic structure that reflects local relationships within the data. We examine how various $k$ values influence classification performance. By setting $k=\{5, 10, ..., 25\}$, we report the F1 scores of TabGSL in Figure~\ref{fig:hyperp_k}. We can see that no consistent tendency appears in the results. The impact of various $k$ values is quite different across datasets. When $k$ increases, the performance scores of some datasets are improved but the scores of other datasets are decreased. Such results may come from the characteristics of datasets, which can be summarized in the following four points. (a) \textit{Dataset Complexity}: In complex datasets where the relationships between instances are more intricate, a larger $k$ value might be beneficial as it could capture more information and result in a richer initial graph structure. Conversely, for simpler datasets, a smaller $k$ might be sufficient, and increasing $k$ might lead to overfitting. (b) \textit{Noise and Outliers}: If a dataset contains a significant amount of noise or outliers, increasing $k$ might lead to the inclusion of more of these noise points in the initial graph structure, which could negatively impact the performance. Conversely, in relatively clean datasets, increasing $k$ might improve performance by leveraging more relevant information. (c) \textit{Class Imbalance}: If the dataset is more imbalanced, a larger $k$ could potentially include more instances of the majority class, thereby overwhelming the minority class and skewing the learned graph structure. This could lead to a decrease in the GCN's performance in the minority class. (d) \textit{Data Density}: In densely populated feature spaces, a smaller $k$ might be enough to capture the local structure around each instance. However, as $k$ increases, the graph might start to capture unnecessary relations between instances, potentially confusing the GCN. Conversely, in sparsely populated spaces, increasing $k$ might help by capturing more distant but potentially meaningful connections. Analyzing the relationship between the dataset characteristics and the initialization of graph structure learning will be an important topic in the future GNN-based tabular data learning.

\section{Visualization of the Learned Graphs}
\label{apd-graphvis}
Pairs of data instances with the same label are supposed to be connected and therefore have a higher edge weight in the adjacency matrix of the learned graph. We investigate whether the graph learned via our TabGSL can produce such an effect. We visualize the learned graph in two manners. One is the adjacency matrix of the learned graph, and the other is the corresponding graph visualization. For the former, we create a heat map for visualizing the adjacency matrix, in which instances with the same class labels in the ground truth and the prediction outcome are arranged in the x-axis and y-axis. Ideally, instances with the same labels will be highlighted in the diagonal blocks. The cell color represents the edge weight. A cell closer to red or black indicates a higher edge weight, and a color closer to yellow or white indicates a lower edge weight. A $100\%$ white cell suggests the absence of the edge. For the latter, we draw the learned graph structure. A linked instance pair is marked as two colorful points connected by a solid line whose thickness indicates the edge weight. The color of a point indicates its class label. We expect that instances with the same labels are connected to each other in the graph visualization.

    \begin{figure}[!t]
    	\centering
    	\includegraphics[width=1.0\textwidth]{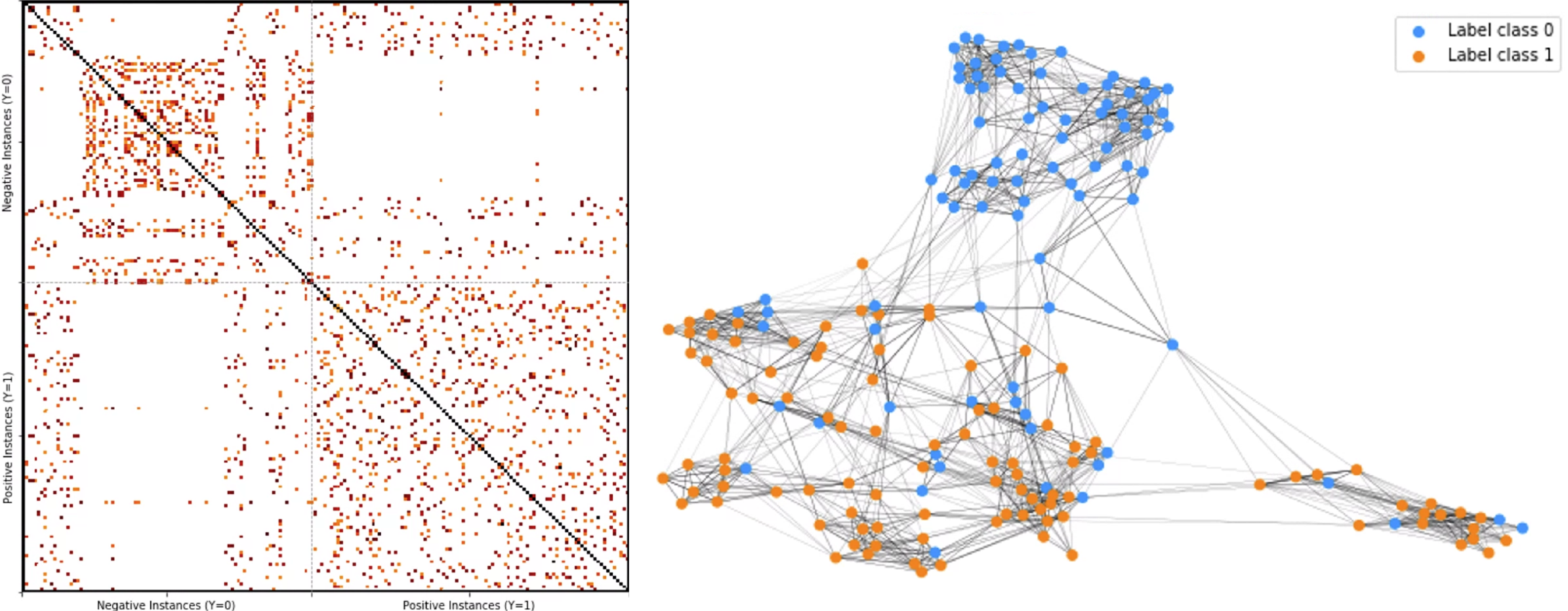}
    	\caption{\textbf{Visualization plots for binary classification dataset $941$.} The adjacency matrix (left) and the graph structure (right) of the graph learned by TabGSL.}
    	\label{fig:vis_941}	
    \end{figure}

    \begin{figure}[!t]
    	\centering
    	\includegraphics[width=1.0\textwidth]{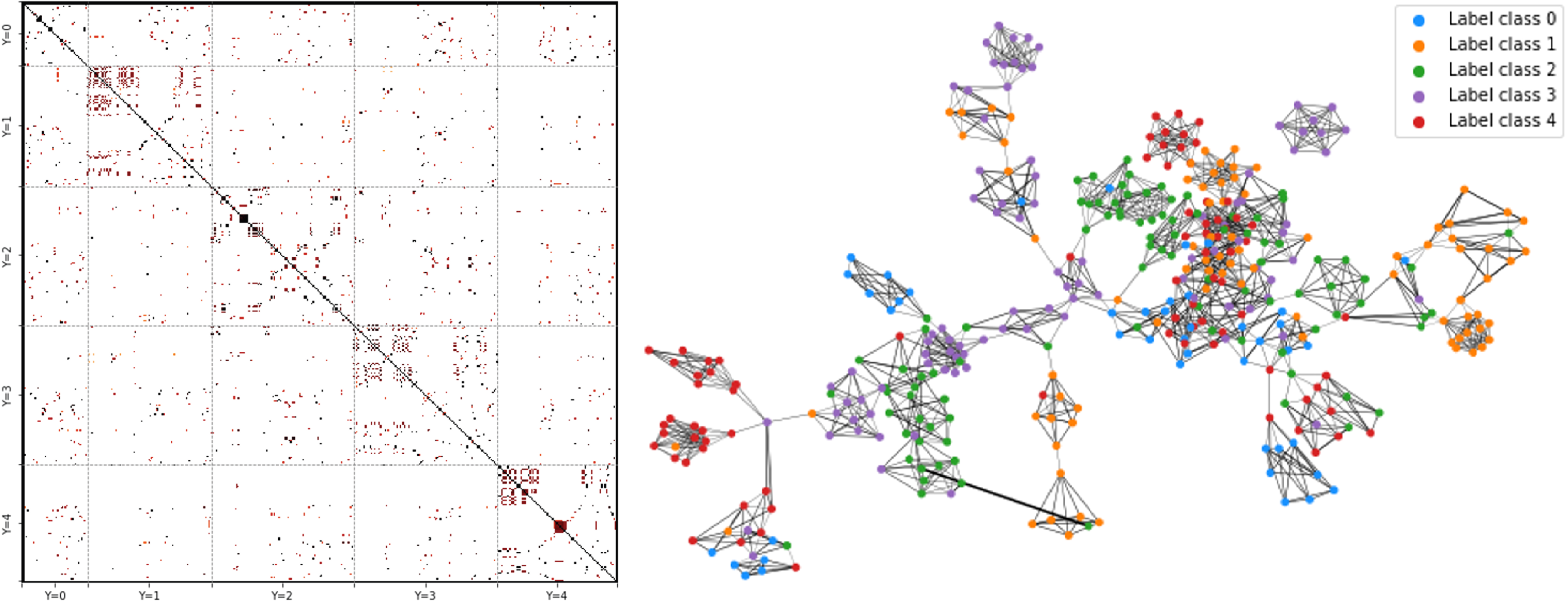}
    	\caption{\textbf{Visualization plots for multi-classification dataset $40663$.} The adjacency matrix (left) and the graph structure (right) of the graph learned by TabGSL.}
    	\label{fig:vis_40663}	
    \end{figure}

By selecting one binary classification dataset $941$ and one multi-class classification dataset $40663$, we provide case studies on the visualizations. The results are exhibited in Figure~\ref{fig:vis_941} and Figure~\ref{fig:vis_40663}, respectively. We can have two main findings. First, the graph learned by the proposed TabGSL does connect instances with the same labels. Although few different-labeled instances are linked, most same-labeled ones tend to connect with each other. Such an effect appears more obvious in dataset 941. For multi-class dataset 40663, although data points of the same label are split into multiple subgraphs, they are still connected in that subgraphs. Second, by looking into the plot of the heat map for the adjacency matrix, there are more red, black, and dark color cells in sub-matrices on the diagonal line, compared to other sub-matrices. This outcome means that instance pairs with the same label tend to obtain higher edge weights when learning the graph structure. These results verify TabGSL's effectiveness in graph structure learning for tabular data. 

Here we aim to further discuss two issues on the learned graph. The first is \textbf{why the connectivity of same-label instances is more obvious in binary classification?} Compared to the multi-class dataset, the binary classification dataset tends to have a simpler structure, with only two class labels to discern. The proposed TabGSL may have an easier time finding and delineating the distinctions between these two labels, resulting in a more obvious pattern in the learned graph. The second is \textbf{why same-label instances form multiple subgraphs in multi-class Classification datasets?} We think there are three possible reasons. (a) \textit{Variations within a Class}: Even within a single class, there could be significant variability in the features. These different ``sub-groups'' within a class could lead to the formation of separate subgraphs. (b) \textit{Non-Linear Separations}: The divisions between classes in a multi-class problem might not be linear or straightforward. If TabGSL is capturing these complex separations, it could result in same-label instances being spread across multiple subgraphs. (c) \textit{Noise and Overfitting}: If the dataset contains a lot of noise, or if TabGSL is overfitting to specific characteristics of the training data, this could also lead to same-label instances being spread across multiple subgraphs. (d) \textit{Feature Complexity}: The complexity of the feature space tends to increase with the number of classes. This might result in instances of the same class appearing dissimilar in the high-dimensional space, leading to the formation of multiple subgraphs.